\documentclass[11pt]{article}

\usepackage[english]{babel}

\usepackage{amsmath}
\usepackage{graphicx}
\usepackage[colorlinks=true, allcolors=blue]{hyperref}
\usepackage{tikz}
\usepackage[utf8]{inputenc}
\usepackage[T1]{fontenc}
\usepackage{amsmath}
\usepackage{amsfonts}
\usepackage{graphicx}
\usepackage{fancyhdr}
\usepackage[margin=1in]{geometry}
\usepackage{hyperref} 
\usepackage{enumitem} 
\usepackage{verbatim}
\usepackage{booktabs} 
\usepackage{braket}
\usepackage{tikz}
\usepackage{natbib}
\usepackage[utf8]{inputenc} 
\usepackage{hyperref}         
\usepackage[official]{eurosym}  

\hypersetup{
    colorlinks=true,
    linkcolor=blue,
    filecolor=magenta,      
    urlcolor=cyan,
    citecolor=blue,
}

\title{\textbf{FD4QC: Application of Classical and Quantum-Hybrid Machine Learning for Financial Fraud Detection} \\ \large A Technical Report}
\author{
  Matteo Cardaioli\thanks{GFT Technologies} \and
  Luca Marangoni\footnotemark[1] \and
  Giada Martini\thanks{Spritzmatter} \and
  Francesco Mazzolin\footnotemark[2] \and
  Luca Pajola\footnotemark[2] \and
  Andrea Ferretto Parodi\footnotemark[1] \and
  Alessandra Saitta\footnotemark[1] \and
  Maria Chiara Vernillo\footnotemark[1]
}

\begin{document}
\maketitle

\makeatletter
\def\ps@myheadings{%
  \def\@oddfoot{\hfil\thepage\hfil}%
\def\@evenfoot{\hfil\thepage\hfil}%
  \def\@oddhead{\hfil Technical Report\hfil}%
  \def\@evenhead{\hfil Technical Report\hfil}%
}
\pagestyle{myheadings}
\makeatother

\begin{abstract}
The increasing complexity and volume of financial transactions pose significant challenges to traditional fraud detection systems. This technical report investigates and compares the efficacy of classical, quantum, and quantum-hybrid machine learning models for the binary classification of fraudulent financial activities.

As of our methodology, first, we develop a comprehensive behavioural feature engineering framework to transform raw transactional data into a rich, descriptive feature set. Second, we implement and evaluate a range of models on the IBM Anti-Money Laundering (AML) dataset. The classical baseline models include \textit{Logistic Regression}, \textit{Decision Tree}, \textit{Random Forest}, and \textit{XGBoost}. These are compared against three hybrid classic quantum algorithms architectures: a \textit{Quantum Support Vector Machine} (\textbf{QSVM}), a \textit{Variational Quantum Classifier} (\textbf{VQC}), and a \textit{Hybrid Quantum Neural Network} (\textbf{HQNN}).

Furthermore, we propose Fraud Detection for Quantum Computing (\textbf{FD4QC}), a practical, API-driven system architecture designed for real-world deployment, featuring a ``classical-first, quantum-enhanced'' philosophy with robust fallback mechanisms.

Our results demonstrate that classical tree-based models, particularly \textit{Random Forest}, significantly outperform the quantum counterparts in the current setup, achieving high accuracy (\(97.34\%\)) and F-measure (\(86.95\%\)). Among the quantum models, \textbf{QSVM} shows the most promise, delivering high precision (\(77.15\%\)) and a low false-positive rate (\(1.36\%\)), albeit with lower recall and significant computational overhead.

This report provides a benchmark for a real-world financial application, highlights the current limitations of quantum machine learning in this domain, and outlines promising directions for future research.

\end{abstract}
\newpage
\section{Introduction} 
Financial institutions face significant financial and reputational risks from fraudulent activities, making them prime targets for advanced detection systems. In the European Economic Area (EEA), fraud losses across major payment instruments totaled \euro4.3 billion in 2022, with an additional \euro2.0 billion reported in the first half of 2023 alone~\cite{ecb_eba_fraud_2024}. For card payments, the fraud rate was 0.031\% of the total transaction value during the first half of 2023, equivalent to 3.1 cents for every \euro100 transacted. The threat is amplified in cross-border transactions; fraud rates for card payments were ten times higher when the counterpart was located outside the EEA, where the application of Strong Customer Authentication (SCA) is not legally required~\cite{ecb_eba_fraud_2024}. Beyond direct financial costs, fraud inflicts substantial reputational damage and erodes customer trust. A recent industry analysis highlights that over 30\% of fraud victims leave their financial institution, underscoring the critical importance of robust security~\cite{aci_trust_2025}. These fraudulent activities are often enabled by sophisticated techniques such as social engineering and phishing, which lead to stolen card details and manipulated credit transfers. 
\par
The inefficiencies of current detection systems, often marked by high false-positive rates, further highlight the need for technological innovation. Financial fraud is a pervasive and evolving threat, requiring the continuous development of sophisticated detection methodologies. While classical Machine Learning (ML) has been the cornerstone of fraud detection systems for years, the escalating complexity of fraudulent schemes and the sheer volume of data are pushing the boundaries of these approaches. Concurrently, the nascent field of Quantum Machine Learning (QML) offers intriguing possibilities, leveraging quantum phenomena like superposition and entanglement to unlock potentially new computational paradigms for complex pattern recognition. 
\paragraph{Purpose and Scope.}
This research provides a comparative analysis of classical and quantum-hybrid ML models for the binary classification of financial transactions. We develop, implement, and evaluate three distinct quantum-inspired architectures: the \textit{Quantum Support Vector Machine (QSVM)}, the \textit{Variational Quantum Classifier (VQC)}, and a \textit{Hybrid Quantum Neural Network (HQNN)}. These are benchmarked against robust classical models. A core contribution of this work is an exploration of behavioural feature engineering, which is crucial for transforming raw data into meaningful inputs. Beyond model development, we address the practical deployment of these technologies through the design of the \textbf{FD4QC} service, an API conceptualised to integrate advanced fraud detection models into operational environments. This technical report synthesizes our findings, reflects on current limitations, and proposes future directions for for future research and experimental improvements.
\section{Methodology}

Our experimental methodology includes data selection and behavioural feature engineering, the implementation of both classical and quantum models, and the design of a deployable system architecture.

\subsection{Dataset and Behavioural Feature Engineering}

Our analysis is based on the synthetic IBM Transactions for Anti-Money Laundering (AML) dataset~\cite{altman2023realistic, egressy2024provably}, which simulates realistic transactional behaviour and suspicious activities  realistic transactional behavior and suspicious activities within a network of customers and financial institutions.\footnote{\url{https://www.kaggle.com/datasets/ealtman2019/ibm-transactions-for-anti-money-laundering-aml}}. This synthetic dataset was generated by IBM to support research and development in the field of financial crime detection, particularly for Anti-Money Laundering tasks. 

To enhance model performance, we engineered a comprehensive set of behavioural features designed to capture the historical patterns of sender and receiver accounts.\footnote{We defined behavioral feature engineering following the idea proposed in~\cite{pajola2023novel}.} These features, calculated based on past activity, can be grouped into several categories:

\begin{itemize}
    \item \textbf{Statistical Features by Account Role:} Mean, standard deviation, max, and min of transaction amounts for sender and receiver accounts, considering past sending and receiving activities separately.
    
    \item \textbf{Temporal Dynamics:} Time elapsed since the last transaction for each account, including standard and \textit{Exponentially Weighted Moving Averages (EWMA)} to smooth patterns.
    
    \item \textbf{Categorical and Contextual Features:} One-hot encoded features for payment currency and format, alongside contextual flags like \texttt{Same\_Bank} or \texttt{is\_self\_loop}.
    
    \item \textbf{Behavioural Change Features:} Indicators that detect deviations from an account's typical transaction patterns in terms of currency or payment format.
    
    \item \textbf{Pairwise Features:} Statistics capturing the specific transactional history between a sender-receiver pair, including a novel \texttt{Pair\_Equilibrium} metric to measure the balance of the relationship.
\end{itemize}

\subsection{Classical Baseline Models}

To provide a robust reference, we evaluated four widely used supervised learning algorithms:

\begin{itemize}
    \item \textbf{Logistic Regression (LR):} a linear model for binary classification.
    
    \item \textbf{Decision Tree (DT):} a non-linear model based on hierarchical, axis-aligned splits.
    
    \item \textbf{Random Forest (RF):} an ensemble method of decision trees to reduce variance.
    
    \item \textbf{XGBoost (XGB):} an efficient, scalable implementation of gradient-boosted decision trees.
\end{itemize}

\subsection{Quantum and Hybrid Models}

We explored three quantum-classical hybrid architectures that integrate quantum computing with classical optimization techniques. These algorithms have been implemented using \textit{PennyLane} and its integrations with \textit{Scikit-learn}\cite{pedregosa2011scikit} and \textit{PyTorch} \cite{paszke2019pytorch}.

\begin{enumerate}
    \item \textbf{Quantum Support Vector Machine (QSVM).} It extends the classical SVM by using a quantum circuit to compute the kernel matrix. Data points are encoded into quantum states, and their inner product in the Hilbert space (an abstract vector space like Cartesian space but possibly infinite-dimensional, fundamental to quantum mechanics) defines the kernel, potentially revealing complex correlations intractable for classical kernels.

    \item \textbf{Variational Quantum Classifier (VQC).} A hybrid Quantum Machine Learning (QML) model consisting of three main components:
    \begin{itemize}
        \item \textit{Classical Data to Quantum States Encoder.} A quantum feature map is used to encode classical input data into quantum states using a quantum circuit. This step is crucial to enable the quantum model to process classical data.

        \item \textit{Parametrised Quantum Circuit (Ansatz).} A quantum circuit that defines the model architecture and contains trainable quantum gate parameters, which determine how the quantum state evolves \cite{haug2021capacity}.

        \item \textit{Classical Optimization Algorithm.} Classical machine learning techniques, such as gradient descent, are used to iteratively update the parameters of the Ansatz.
    \end{itemize}

    \item \textbf{Hybrid Quantum Neural Network (HQNN).} It integrates a variational quantum circuit as a hidden layer within a classical deep neural network. A classical encoder compresses the input features, which are then processed by the quantum layer. The results are fed back into a classical classifier for the final prediction. This architecture follows the paradigm outlined in~\cite{bergholm2022pennylane,bischof2025hybrid}.
\end{enumerate}

\subsection{System Architecture: The FD4QC Service}

To address practical deployment challenges, we designed \textbf{FD4QC}, a stateless RESTful API. Its ``classical-first, quantum-enhanced'' philosophy ensures operational robustness by using proven classical models as the backbone, while allowing for the controlled, gradual integration and A/B testing of experimental quantum models.

The system includes a lightweight router for model selection and a transparent fallback mechanism: if a quantum backend is unavailable, the request is automatically rerouted to a classical surrogate, ensuring service continuity. The API response explicitly flags whether the prediction was generated by a ``classical'' or ``quantum'' engine, ensuring auditability.

\section{Experimental Setup and Results}
All models were trained and evaluated on a reduced and undersampled version of the IBM dataset to facilitate experimentation, maintaining a class ratio of 9: 1 (non-suspicious to suspicious). The performance of the models was assessed on a holdout test set using \textbf{Accuracy}, \textbf{F-measure}, \textbf{Precision}, \textbf{Recall}, and the \textbf{False Positive Rate (FPR)}.

Although real quantum hardware was not used in this study, the code implemented is compatible with quantum devices and can be executed on actual hardware via the \textit{Pennylane} \cite{bergholm2022pennylane} interface for quantum computing platforms.

The comparative results are summarized in Table 1.

\begin{table}[ht]
\centering
\label{tab:results_bench}
\begin{tabular}{|l|c|c|c|c|c|}
\hline
\textbf{Algorithm} & \textbf{Accuracy} & \textbf{F-measure} & \textbf{Precision} & \textbf{Recall} & \textbf{FPR} \\
\hline
\multicolumn{6}{|c|}{\textbf{Classical Models}} \\
\hline
Logistic Regression        & 0.8588 & 0.1241 & 0.1634 & 0.1000 & 0.0569 \\
Decision Tree              & 0.9652 & 0.8374 & 0.7860 & 0.8960 & 0.0271 \\
Random Forest              & 0.9734 & 0.8695 & 0.8536 & 0.8860 & 0.0169 \\
XGBoost                    & 0.9698 & 0.8558 & 0.8190 & 0.8960 & 0.0220 \\
\hline
\multicolumn{6}{|c|}{\textbf{Quantum Models}} \\
\hline
VQC -- 1 layer -- 4 qubits     & 0.5990 & 0.2128 & 0.1324 & 0.5420 & 0.3947 \\
VQC -- 2 layers -- 4 qubits    & 0.5024 & 0.1566 & 0.0943 & 0.4620 & 0.4931 \\
HQNN -- 1L* -- 4 qubits        & 0.9000 & 0.0000 & 0.0000 & 0.0000 & 0.0000 \\
HQNN -- 2L* -- 4 qubits        & 0.8448 & 0.0827 & 0.1012 & 0.0700 & 0.0691 \\
QSVC -- 2 qubits               & 0.9272 & 0.5297 & 0.7482 & 0.4100 & 0.0153 \\
QSVC -- 4 qubits               & 0.9290 & 0.5372 & 0.7715 & 0.4120 & 0.0136 \\
\hline
\end{tabular}
\caption{Performance metrics of classical and quantum-inspired models on the test set. L* indicates the number of quantum layers.}
\end{table}

\subsection{Analysis of Results}

\begin{itemize}
    \item \textbf{Classical Baselines} 
    \newline Tree-based ensemble models (\textit{Random Forest} and \textit{XGBoost}) demonstrate clear superiority, achieving excellent balance across all metrics. \textit{Random Forest} shows the best overall performance, with high F-measure and a low FPR, which is critical for minimizing false alarms in an operational setting.
        
    \item \textbf{Quantum-Inspired Models}
    
    \begin{itemize}
        \item \textit{VQC} and \textit{HQNN} models performed poorly. Despite relatively high accuracy scores (attributable to the class imbalance), their F-measure and Recall were near-zero, indicating a failure to identify the positive (fraudulent) class. Their behaviour suggests a bias towards the majority class.
    
        \item \textit{QSVC} emerged as the most viable quantum model. The 4-qubit configuration achieved a respectable accuracy (92.9\%) and F-measure (53.72\%). Notably, it delivered high precision (77.15\%) and a very low FPR (1.36\%), comparable to the best classical models. However, its Recall (41.20\%) is substantially lower than classical baselines, and its training/inference times were significantly longer, posing a practical challenge.
    \end{itemize}
    
\end{itemize}

\section{Discussion}

The experimental results clearly highlight that for this fraud detection task, well-established classical ensemble methods, powered by domain-specific feature engineering, remain the preferred choice for reliable and effective detection. The superior performance of \textit{Random Forest} and \textit{XGBoost} suggests that the rich, engineered features provide a sufficiently expressive data representation that these models can effectively exploit.

The underperformance of the quantum models can be attributed to several factors, including algorithmic immaturity, challenges in training variational circuits, and the possibility that the dataset, even with complex features, does not possess the specific structure that would unlock a quantum advantage.

Despite this, the performance of the \textit{QSVC} is noteworthy. Its ability to achieve high precision and a low FPR suggests a potential niche in settings where the cost of a false positive is extremely high, and a lower detection rate (recall) is, in turn, an acceptable trade-off. This finding warrants further investigation into quantum kernel methods.

Several promising research directions could improve the viability of QML for this application:

\begin{itemize}
    \item \textbf{Advanced Circuit Design: } Exploring tailored ansätze and adaptive feature maps to enhance model expressiveness without incurring training instabilities.

    \item \textbf{Hybrid Architectures: }Combining temporal feature extraction (e.g., with LSTMs) with variational quantum circuits to better leverage the strengths of both paradigms \cite{takaki2021learning}.

    \item \textbf{Data Encoding \& Feature Selection}: Utilizing techniques like quantum autoencoders \cite{romero2017quantum} for feature compression or metaheuristic methods for feature selection to prepare high-dimensional data for quantum processing.
\end{itemize}

\section{Conclusion}

This study provides a benchmark of classical, quantum-hybrid, and quantum machine learning models for financial fraud detection. Our findings indicate that, at present, quantum AI is not mature enough to outperform traditional algorithms in this practical, high-stakes domain. Classical ensemble methods demonstrate robust, superior performance, making them the preferred choice for operational deployment.

However, our exploration has yielded valuable insights. The \textit{QSVC} model showed notable potential in achieving high precision, and the conceptual \textbf{FD4QC} architecture offers a pragmatic roadmap for integrating future quantum capabilities into financial security systems. Continued investigation into quantum kernel methods, advanced hybrid architectures, and sophisticated data encoding strategies is essential for unlocking the future potential of quantum computing in the financial sector.

\section{Acknowledgments}
This work was supported by the project FD4CQ – Fraud Detection con Computer Quantistici, funded under the Italian PNRR initiative, Mission 4 “Education and Research” – Component 2 “From Research to Business”, Investment Line 1.4, and financed by the European Union – NextGenerationEU. The project is part of ICSC – Spoke 10 "Quantum Computing".

\bibliographystyle{plain}
\bibliography{references}
\end{document}